%% file: main.tex
\documentclass{pprai}

\usepackage[utf8]{inputenc}
\usepackage[T1]{fontenc}
\usepackage{graphicx}
\usepackage{amsthm}
\usepackage{txfonts}
\usepackage{url}
\usepackage{algpseudocode}
\usepackage{multirow}
\usepackage[colorinlistoftodos]{todonotes}

\title{3D Reconstruction of non-visible surfaces of objects from a Single Depth View -- Comparative Study}
\headtitle{3D Reconstruction of non-visible surfaces of objects\dots}


\author{Rafa{\l} Staszak$^{1[0000-0002-5235-4201]}$, Piotr Micha{\l}ek$^{1[0009-0003-9139-4665]}$, \\Jakub Chudzi\'nski$^{1[0000-0001-8228-0197]}$, Marek Kopicki$^{1[0000-0002-0769-0556]}$, \\Dominik Belter$^{1[0000-0003-3002-9747]}$}
\headauthor{R. Staszak, P. Micha{\l}ek, J. Chudzi\'nski, M. Kopicki, D. Belter}
\affiliation{%
  $^1$Poznan University of Technology\\
  Institute of Robotics and Machine Intelligence\\
  ul. Piotrowo 3A, 60-965 Pozna\'{n}, Poland\\
 name.surname@put.poznan.pl}


\graphicspath{ {figures/} }

\begin{document}
\maketitle

\begin{abstract}
Scene and object reconstruction is an important problem in robotics, in particular in planning collision-free trajectories or in object manipulation. This paper compares two strategies for the reconstruction of non-visible parts of the object surface from a single RGB-D camera view. The first method, named DeepSDF predicts the Signed Distance Transform to the object surface for a given point in 3D space. The second method, named MirrorNet reconstructs the occluded objects' parts by generating images from the other side of the observed object. Experiments performed with objects from the ShapeNet dataset, show that the view-dependent MirrorNet is faster and has smaller reconstruction errors in most categories.
\end{abstract}

\section{Introduction}

\begin{figure}[ht] 
\centering
\includegraphics[width=0.95\textwidth]{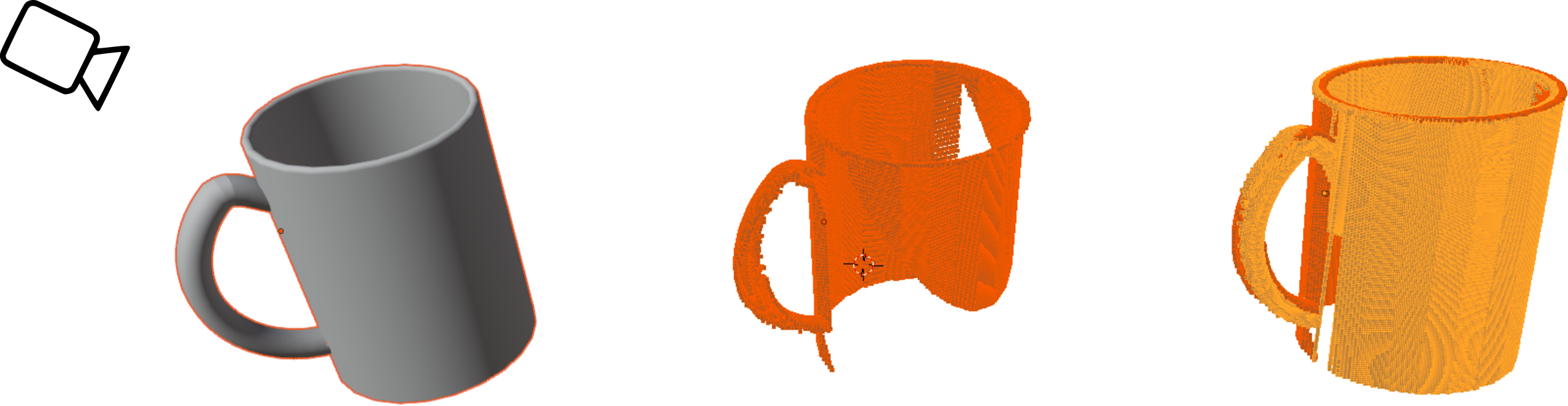}
\put(-358,85){a} \put(-190,82){b} \put(-74,82){c}
\caption{Example application scenario of the objects reconstruction system: the robot observes a 3D object from a single view (a). The incomplete model of the object (point cloud) (b) is provided to the input of the neural network to obtain a full model of the object (c).}
\label{fig:intro}
\end{figure}

Robots observing the scene utilize onboard RGB-D cameras to collect information about the shape of the objects. However, the full geometry of the scene cannot be registered from a single view due to occlusions. Some methods for grasping objects deal with incomplete data and perform well even though the full 3D model is unknown~\cite{Kopicki2019}. In this research, we are focused on the solutions that directly reconstruct the entities on the scene. The example object reconstruction scenario is presented in Fig.~\ref{fig:intro}. The object reconstruction method that can be applied in robotics should be capable of reconstructing a full model of an entity observed from a single RGB-D camera view. The features extracted by the considered solutions are potentially valuable for other robotics tasks e.g. grasping~\cite{Weng2022}.

Multiple scene reconstruction techniques utilize 3D grids~\cite{Choy2016,Popov2020} but these methods suffer from resource consumption growth when the resolution of the model is increased. Recently the Neural Scene Representation model based on Radiance Fields (NeRF) has been proposed~\cite{mildenhall2020nerf}. This solution has superior accuracy but is designed for generating images from various viewpoints for static scenes. This property limits the possible applications in robotics. In contrast, DeepSDF~\cite{Park2019} can be used to reconstruct various objects from a single view and represent them as a Signed Distance Transform (SDF). Other methods are designed to generate depth images of the observed objects from various viewpoints~\cite{Staszak2022}. Thus, the obtained images are used to reconstruct a full 3D model of the entity. In this paper, we compare the view-dependent approach based on image generation named MirrorNet~\cite{Staszak2022} with the SDF-based neural representation operating directly in the continuous 3D space~\cite{Park2019}.

\section{Comparison between DeepSDF and MirrorNet}

In this paper, we compare two representative approaches for scene reconstruction. The DeepSDF method~\cite{Park2019} utilizes a fully-connected neural network to predict the Signed Distance Transform to the surface of the object for the given point in the 3D space. During the inference, the latent space, that describes the object, is estimated from the partial view. Then, the object is reconstructed by sampling the 3D space and generating new points. The second method proposed in~\cite{Staszak2022} reconstructs the occluded parts of the objects by generating images from the other side of the observed object. It utilizes the depth image from the given position of the camera and the depth image obtained by projecting the input point cloud to the virtual position of the camera. In this case, a Convolutional Neural Network is used to generate depth images.

Both methods are designed to operate in slightly different conditions. DeepSDF requires depth observations in the canonical shape frame of reference. The MirrorNet does not have this limitation but directly utilizes noisy depth camera images from the real robot~\cite{Staszak2022}. To compare both methods in the same condition, we select 6 representative categories of graspable entities from the ShapeNet dataset~\cite{Chang2015}. For each category, we selected 30 instances of objects to prepare the training datasets. To train the DeepSDF, we utilize a 3D mesh model of the objects located in the global frame and scaled to fit the unit sphere. To generate training data for the MirrorNet, we collect images generated for random positions around the object. For testing, we use randomly generated views of objects for another 10 instances of objects from the categories that were used for training.
 
\section{Results}

\begin{table}[t]
\caption{Comparison of the reconstructed results obtained for the view-dependent model (MirrorNet)~\cite{Staszak2022} and DeepSDF~\cite{Park2019}.}\label{tab:results}
\centering
\begin{tabular}{|l|ccccccc|}
\hline
\textbf{method} & \textbf{metric} & \textbf{bottle} & \textbf{can} & \textbf{helmet} & \textbf{jar} & \textbf{laptop} & \textbf{mug} \\ 
\hline\hline
\multirow{3}{*}{MirrorNet} & $d_{\rm C}$ [m] & {\bf 0.06836} & {\bf 0.1402} & 0.1156 & {\bf 0.0749} & 0.1967 & 0.1843\\ 
 & $d_{\rm H}$ [m] & {\bf 0.0997} & {\bf 0.1647} & {\bf 0.1673} & {\bf 0.1257} & {\bf 0.2606} & {\bf 0.2176}\\ 
 \multirow{3}{*}{DeepSDF} & $d_{\rm C}$ [m] & 0.09764 & 0.1486 & {\bf 0.0724} & 0.0803 & {\bf 0.1437} & {\bf 0.1125}\\ 
 & $d_{\rm H}$ [m] & 0.3143 & 0.4162 & 0.3098 & 0.3071 & 0.4062 & 0.3596\\ 
\hline
\end{tabular}
\end{table}

The obtained results are presented in Tab.~\ref{tab:results}. We utilize Chamfer $d_C$ and Hausdorff $d_H$ distances~\cite{Lebrat2021} to quantitatively evaluate the reconstruction results in 3D space. Both systems return similar results regarding the Chamfer distance $d_C$. Despite the visually better reconstruction of the objects by the MirrorNet, the obtained 3D model does not cover the whole surface which results in worse numerical results for the Chamfer distance. When the Hausdorff distance $d_H$ is compared the model returned by the MirrorNet is 2-3 times better than the model given by the DeepSDF. 

\begin{figure}[ht] 
\centering
\includegraphics[width=0.99\textwidth]{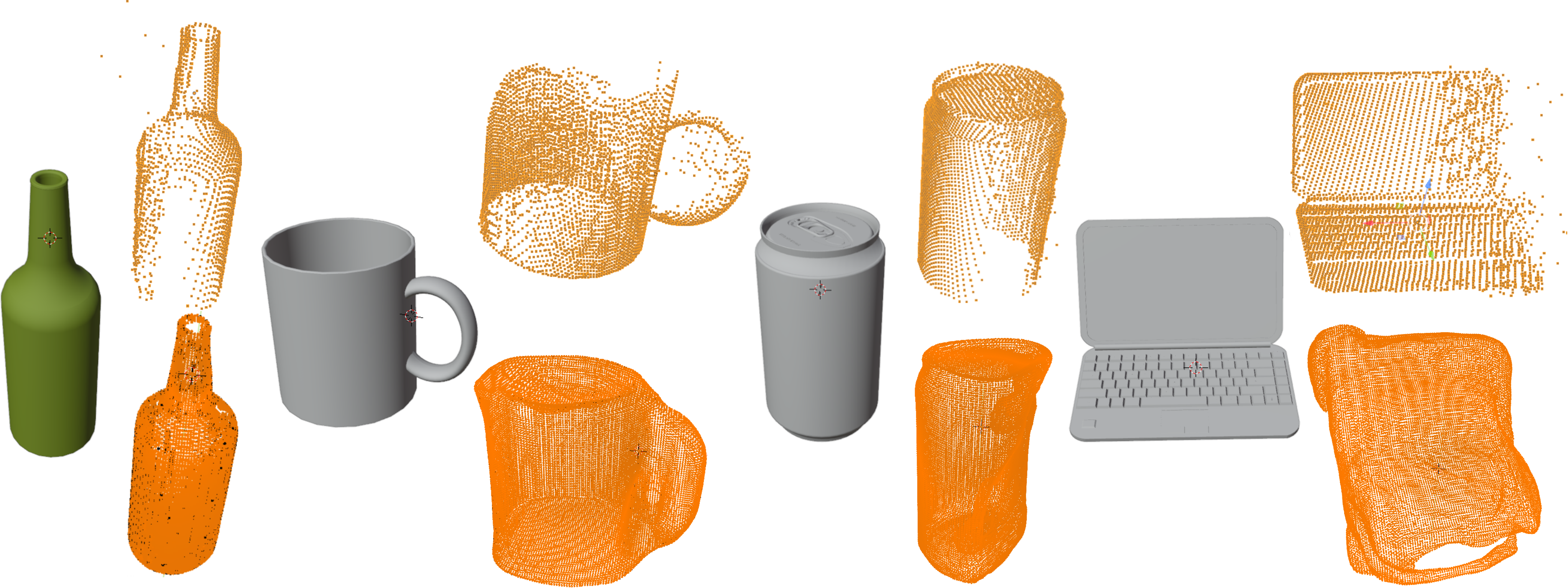}
\caption{Example reconstruction results (point clouds) obtained for the view-dependent MirrorNet (top row) and DeepSDF (bottom row) compared to the ground truth models: bottle, mug, can, and laptop.}
\label{fig:results_MirrorNet}
\end{figure} 

The example reconstruction results obtained from the MirrorNet and DeepSDF are presented in Fig.~\ref{fig:results_MirrorNet}. The first visible difference is the number of points generated by the algorithms which is much smaller for the MirrorNet. The DeepSDF can generate higher number of points but the generation time significantly increases to 15.04 seconds per object (using GPU RTX 3060). For better results with DeepSDF, we increased the number of points drawn in preprocessing from 250,000 to 5 million and introduced a threshold for negative SDF values. The introduction of the threshold was necessary because the algorithm had problems calculating SDF values for an incomplete object mesh from a single view. DeepSDF fails to recover the handle of the mug and the laptop. Also, DeepSDF does not preserve the rounded shape of the can. The view-dependent MirrorNet is fully convolutional so it generates images in about 22 milliseconds. This method preserves the shape of the objects but some surfaces of the objects are not reconstructed because they are not observed from the input and generated camera view. Also, MirrorNet generates random sparse points between the reconstructed surface and the camera pose. However, these points can be easily removed using voxel-based filters.

\section{Conclusions}
In this paper, we compare two neural network-based models that reconstruct the model of the objects from a single camera image. We have chosen the MirrorNet which generates the depth image of the object from the opposite pose of the camera and DeepSDF which operates directly in the 3D space. Our experiments show that the view-dependent approach returns more accurate reconstruction results. Moreover, the view-dependent approach is significantly faster than DeepSDF (22~ms for inference using MirrorNet and 15000~ms for DeepSDF) which requires optimization and 3D space sampling during the inference.

In the future, we are going to extract the features from the view-dependent models of the objects and use them for efficient grasping and manipulating objects from a single camera view.


\section*{Acknowledgment}
\noindent
This research is part of the project No. 2021/43/P/ST6/01921 co-funded by the National Science Centre and the European Union Framework Programme for Research and Innovation Horizon 2020 under the Marie Skłodowska-Curie grant agreement No. 945339.
R. Staszak and D. Belter were supported by the National Science Centre, Poland, under research project no UMO-2019/35/D/ST6/03959.


\bibliography{pprai}
\bibliographystyle{pprai}

\end{document}